\documentclass[letterpaper, 10 pt, conference]{ieeeconf}  

\IEEEoverridecommandlockouts                              

\pdfoutput=1

\usepackage{graphics} 
\usepackage{epsfig} 
\usepackage{times} 
\usepackage{amsmath} 
\usepackage{amssymb}  
\usepackage{subcaption}

\usepackage{graphics} 
\usepackage{epsfig} 
\usepackage{times} 
\usepackage{amsmath} 

\usepackage{amssymb}  

\usepackage{booktabs}
\usepackage{dblfloatfix}
\usepackage{graphicx} 
\usepackage{caption}

\let\labelindent\relax
\usepackage{enumitem}
\setlist[itemize,1]{leftmargin=\dimexpr 26pt-.2in}
\setlist[enumerate,1]{leftmargin=\dimexpr 26pt-.2in}
\usepackage{cuted} 
\usepackage{capt-of}

\usepackage{subcaption}
\usepackage{bbding}
\usepackage{color}
\usepackage{float} 
\usepackage{algorithm}
\usepackage[noEnd=True]{algpseudocodex}

\makeatletter
\let\NAT@parse\undefined
\makeatother
\usepackage{hyperref}  

\newcommand{\kw}[1]{\textbf{#1}}

\newcommand{\True}[0]{\kw{True}}

\newcommand{\pddl}[1]{{\texttt{#1}}} 
\newcommand{\pddlsmall}[1]{{\small \texttt{#1}}} 

\newcommand{\tamp}[0]{{TAMP}}
\newcommand{\dimsam}{{\sc DiMSam}}

\usepackage{algorithm}
\usepackage[export]{adjustbox}
\algnewcommand\algorithmicdeclare{\textbf{Assume:}}
\algnewcommand\Declare{\item[\algorithmicdeclare]}

\usepackage{listings}
\usepackage{bm}
\usepackage[colorinlistoftodos, textwidth=3.5cm]{todonotes}

\title{\LARGE \bf
DiMSam: \textit{Di}ffusion \textit{M}odels as \textit{Sam}plers for\\
Task and Motion Planning 
under Partial Observability
}

\author{Xiaolin Fang$^{1}$, Caelan Reed Garrett$^{2}$, Clemens Eppner$^{2}$, \\
Tom\'as Lozano-P\'erez$^{1}$, Leslie Pack Kaelbling$^{1}$, Dieter Fox$^{2}$
\thanks{$^{1}$MIT CSAIL
        {\tt\small \{xiaolinf,tlp,lpk\}@csail.mit.edu}}%
\thanks{$^{2}$NVIDIA
        {\tt\small \{cgarrett,ceppner,dieterf\}@nvidia.com}}%
}

\begin{document}

\maketitle
\thispagestyle{empty}
\pagestyle{empty}


\begin{abstract}
Generative models such as diffusion models, excel at capturing high-dimensional distributions with diverse input modalities, e.g. robot trajectories, but are less effective at multi-step constraint reasoning.
Task and Motion Planning (TAMP) approaches are suited for planning multi-step autonomous robot manipulation.
However, it can be difficult to apply them to domains where the environment and its dynamics are not fully known. 
We propose to overcome these limitations by composing diffusion models using a \tamp{} system. We use the learned components for constraints and samplers that are difficult to engineer in the planning model, and use a TAMP solver to search for the task plan with constraint-satisfying action parameter values.
To tractably make predictions for unseen objects in the environment, we define the learned samplers and \tamp{} operators on learned latent embedding of changing object states.
We evaluate our approach in a simulated articulated object manipulation domain and show how the combination of classical TAMP, generative modeling, and latent embedding enables multi-step constraint-based reasoning. We also apply the learned sampler in the real world. Website: \href{https://sites.google.com/view/dimsam-tamp}{https://sites.google.com/view/dimsam-tamp}.
\end{abstract}


\section{INTRODUCTION} \label{sec:intro}

Autonomous robot manipulation in real-world environments is challenging due to large action spaces, long periods of autonomy, the need for contact-rich interaction, and the presence of never-before-seen objects.
Although it is in principle possible to learn direct policies for manipulation through imitation or reinforcement learning, these methods generally have particular difficulty as the action space dimensionality and behavior horizon increase. 
Nonetheless, we have seen prominent progress in generative modeling, that demonstrates its suitability for learning relatively short-horizon robot action distributions. Task and Motion Planning (\tamp{})~\cite{Garrett2021} approaches have an advantage on long-horizon tasks, because they perform model-based reasoning to search over possible futures. In this work, our goal is to compose learned models using a \tamp{} framework for solving multi-step robot manipulation problems.

Solving a long-horizon robot planning problem involves determining an appropriate sequence of action instances (the type of action and the objects it is applied to), and finding the values of the continuous-valued parameters for each of the actions. This process can be characterized as constructing and solving a constraint-satisfaction problem (CSP). The high level task plan implies a set of constraints, which defines a CSP. For example, a task plan involving opening the microwave and then stowing an object implies a constraint that the microwave must be sufficiently opened to stow the object. The planner often needs to alternate between the process of constructing and solving CSPs, as some of the CSPs constructed may be unsatisfiable, such as when the robot seeks an inverse kinematics (IK) solution to an unreachable pose. To effectively address this problem, we need to reason about constraints and efficiently find values to solve the CSP.

A \tamp{} solver is adept at such reasoning procedures. However, finding the continuous action parameter values is challenging. For example, to place an object into a closed microwave, the action parameters include an opened configuration of the microwave, a robot trajectory that opens the microwave to that configuration, a placement pose of the object inside the microwave, and a robot trajectory to place the object that satisfies collision constraints. The interdependence among the constraints makes finding satisfying values difficult. We seek to improve sampling-based \tamp{} by learning to generate constraint-satisfying values.

\begin{figure}[t]
    \centering
    \includegraphics[trim={0cm 0 8.5cm 0}, clip, width=1.0\linewidth]{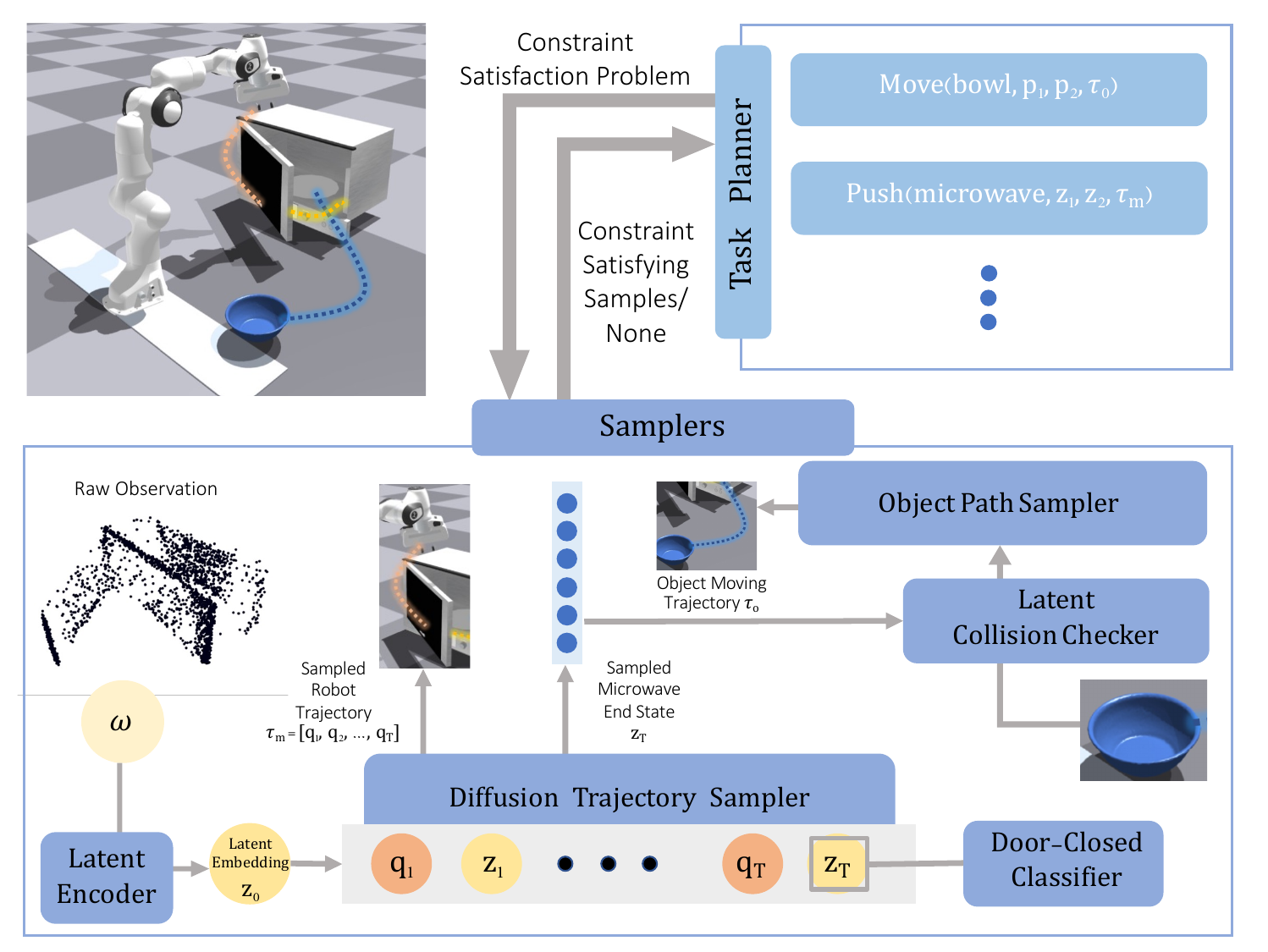}
    \caption{\small{\dimsam{} 
    composes diffusion models using a \tamp{} system
    towards solving multi-step manipulation problems. The planner searches for the task plan, while learned diffusion samplers find constraint-satisfying continuous values. The bottom shows a sampling procedure for finding a microwave door closing trajectory and a 
    collision-free object stowing trajectory. A diffusion model samples a trajectory of latent microwave states $z_1, ..., z_T$ and robot configurations $q_1, ..., q_T$ that reaches a door-closed state $z_T$.
    } } 
    \label{fig:teaser}
\end{figure}

We use learned generative models to represent distributions over continuous state and action parameters. Generative models are trained on a distribution $p(X)$ of possibly complicated variables $X$ (such as images, point clouds, or trajectories), and can be queried to produce samples $x \sim p(X)$ at test time. Specifically, we use {\em diffusion models}~\cite{sohl2015deep, ho2020denoising}, and issue conditional sample queries during inference.

The flexibility in generative models is also valuable in cases of increased partial observability. Learned generative samplers can be conditioned directly on an image or point cloud. We apply diffusion models to point cloud observations with severe partial observability and show the generalizability through real-world verification. 

We propose to use diffusion models as samplers for \tamp{} under partial observability (\dimsam{}). \dimsam{} (Fig.~\ref{fig:teaser}) leverages both the flexibility of generative models in action parameter sampling and the joint searching and sampling capability of \tamp{} systems. The system searches for feasible task plans, with continuous values suggested by learned samplers. By composing these generative models using the \tamp{} framework, our system can 
solve a variety of multi-step manipulation problems under partial observability.

\textbf{The contributions of this work are as follows:}
\begin{enumerate}
    \item We use diffusion models as a generative representation of TAMP constraints in the form of samplers. These learned models are composed through the \tamp{} framework, enabling simultaneous search over constraint structure and sampling on target distribution.
    \item We define these constraints on a latent embedding of object state, allowing them to be applied to previously unseen articulated objects with no known models.
    \item We showcase our approach on articulated object manipulation tasks that require multi-step reasoning and verify a learned sampler in the real world without finetuning.
\end{enumerate}


\section{Related work}

We build on prior work in \tamp{} and generative modeling.
\tamp{} considers planning in {\em hybrid} spaces where there are both continuous and discrete state and action parameters\cite{toussaint2015logic, Garrett2021,curtis2022long, garrett2020online}.
Traditionally, these models are specified by a human; however, increasingly, aspects of these models, such as constraints~\cite{Mao2022PDSketch}, samplers~\cite{Kim2018Gan, wang2021learning}, predicates~\cite{migimatsu2022symbolic}, and parametric operators~\cite{silverlearning}, have been learned. An earlier work~\cite{curtis2022long} extended \tamp{} methods to manipulation without shape models, with a perception module for state estimation. Our method incorporates learned distributional information into the samplers, and also addresses articulated objects.

There is also related work on learning for other manipulation planning frameworks ~\cite{qureshi2021CoRR, driess2022learning, xu2021deep}, but usually not considering complicated objects or partial observability. Our method uses a generative diffusion model to learn samplers based only on the partially observed state of the world, and solve \tamp{} problems in a hybrid space of learned latent and human-specified states.

Generative models~\cite{sohl2015deep, ho2020denoising} have been used in decision making~\cite{janner2022diffuser, ajay2022conditional} and policy learning~\cite{chi2023diffusionpolicy, mishra2023generative, carvalho2023motion}. Compared to non-generative trajectory modeling methods~\cite{janner2021sequence, chen2021decision, liu2022masked}, they are usually better at capturing the multi-modalities in the training data. It has also been used to learn the distribution of parameters for individual samplers~\cite{yang2023diffusion, mendez-mendez2023embodied, liu2023structdiffusion, simeonov2023rpdiff, urain2023se3diffusion}. 
In this paper, we apply a similar model but focus on how to compose the learned models using a \tamp{} system. Previous work usually assumes a given task plan or a constraint graph. Our goal is to use the learned model \textit{combinatorially} with other samplers to solve multi-step planning tasks, in a framework that {\em jointly searches for satisfiable task plans and samples constraint-satisfying values}.

\section{Sampling-based TAMP}

We are interested in using generative learning to extend the applicability of \tamp{} to partially observed settings where classic engineering approaches can't be directly applied.

\subsection{TAMP Problem Description}

A generic \tamp{} problem $\Pi = \langle {\cal S}, {\cal A}, s_0, S_* \rangle$ can be described by a state-space ${\cal S}$, a set of parameterized actions ${\cal A}$, an initial state $s_0 \in {\cal S}$ and a set of goal states $S_* \subseteq {\cal S}$. 
States are comprised of a set of hybrid variables with values that can change over time.
Each parameterized action $a \in {\cal A}$ takes in a tuple of hybrid parameters $x$ that instantiate the {\em preconditions} and {\em effects} of the action.
The preconditions are constraints that enable {\em action instance} $a(x)$ to be correctly executed in state $s$.
The effects specify changes to the variable values in state $s$ that give rise to subsequent state $s'$ after executing action instance $a(x)$.
Critically, in order for an action instance $a(x)$ to be valid, its parameters $x$ must satisfy a conjunctive set of {\em constraints} $a.\kw{con} = \{c_1, ..., c_n\}$, namely $\bigwedge_{i = 1}^n [c_i(x) = \kw{True}]$.

The objective of planning is to find a {\em plan} $\pi$, a finite sequence of action instances $\pi = [a_1(x_1), ..., a_k(x_k)]$, that when executed from state $s_0$, produces a state $s_* \in S_*$.
Actions in a plan often share parameters due to variables persisting in the state.
Thus, the parameters across a valid plan $\bigcup_{i = 1}^kx_i$ must jointly satisfy the corresponding set of action constraints $C_\pi = \bigcup_{i = 1}^k a_i.\kw{con}$.
Solving a \tamp{} problem requires simultaneously identifying a sequence of parameterized actions $[a_1, ..., a_k]$ along with parameter values $[x_1, ..., x_k]$ that satisfy their constraints. 

\subsection{Conditional Samplers}
\label{subsec:condsampledesiderata}

Given a set of constraints $C$, solving for parameter values that satisfy them is a Hybrid Constraint Satisfaction Problem (H-CSP)~\cite{Garrett2021}.
These problems are addressed using 
joint {\em optimization}~\cite{toussaint2015logic} and individual {\em sampling}~\cite{garrett2020PDDLStream} techniques.
Joint optimization methods typically treat the ensemble of parameters and constraints as a single mathematical program and solve for satisfying values all at once.

In contrast, individual sampling techniques leverage {\em compositionality} through conditional sampling, where the outputs of a sampler for one constraint, {\it e.g.}, a grasp pose of an object, become the inputs to another, {\it e.g.}, an inverse kinematics~(IK) solver.
A {\em conditional sampler} for a constraint $c$ with $m$ arguments is a function from $\alpha$, a tuple of $k < m$ argument values, to a generator of tuples $\beta_i$. Each $\beta$ is of length $m-k$. When concatenated with the values $\alpha$, the resulting length-$m$ tuple of values $x_i$ satisfies $c$, 
({\it i.e.} $c(\alpha, \beta) = \True$).
We seek samplers that are:
\begin{itemize}
    \item \textbf{Sound:} they only generate values that satisfy the constraint they represent.
    \item \textbf{Diverse:} they generate multiple diverse values that can be filtered with other constraints via rejection sampling. 
    \item \textbf{Compositional:} they can be combined with others to produce joint samples that satisfy multiple constraints. 
\end{itemize}

In the articulated object manipulation domain described in Sec.~\ref{sec:intro},
we require samplers to generate values under certain constraints, such as 1) generating stable placement and grasp poses, 2) checking collisions at state $s$, 3) generating desirable door states (open or closed), and 4) modeling the contact dynamics of doors.
When the world is known, samplers 1-3 can be engineered.
However, engineering sampler 4 accurately is non-trivial due to the contact-rich nature of the interaction.
Moreover, when the world is partially observable and we do not {\it a priori} know the geometry of objects, engineering samplers 1-3 themselves is challenging.

In our approach, we use diffusion models to learn conditional samplers that represent generative models of action constraints. We show that learned samplers can be applied to parameters that describe the latent state of unknown objects, and allow the planner to search for plans in the latent space. 
The diffusion sampling procedure can also be biased towards generating conditional output by having a classifier as guidance. 

\section{Diffusion Models as Learned Samplers} \label{sec:diffusion}

We seek to learn samplers that generate samples $x$ that satisfy constraint $c$.
We require a {\em training dataset} $\mathcal{D}_c=\{x_{1}, ..., x_{N}\}$ of $N$ length-$m$ parameter tuples $x_{i}$ that satisfy constraint $c$, {\it i.e.} $c(x_{i}) = \True$.
Then, we learn an implicit probability distribution $p(x)$ over parameters $x$ that satisfy constraint $c$ using dataset $\mathcal{D}_c$.
Finally, through incorporating condition terms in the sampling process, we turn the unconditional model $p(x)$ into conditional models $p(\beta \mid \alpha)$ that become the basis for conditional constraint samplers.

\subsection{Diffusion Models}

We use diffusion models as learned samplers. Diffusion models~\cite{sohl2015deep, ho2020denoising} are a class of generative models that produce samples $x^{(0)}$ from a learned distribution $p(x)$ by iteratively applying a denoising procedure $p(x^{(t-1)}\mid x^{(t)})$, starting from $x^{(T)}$, a sample from the Gaussian noise distribution. 
The denoising procedure makes transitions according to
{
\begin{equation}\label{eq:reverse}
    p(x^{(t-1)} \mid x^{(t)}) := \mathcal{N}(x^{(t-1)}; \mu_\theta(x^{(t)},t),\Sigma_\theta(x^{(t)},t)).
\end{equation}
}

Here, $\mu_\theta$ and $\Sigma_\theta$ are time-conditional functions with learnable parameter~$\theta$. 
During training, a forward process gradually adds random noise to the original data point $x^{(0)}$. 
The network is trained to predict the noise added on a data point to generate the corrupted data point. Once the network is trained, the model can be used to draw samples from $p(x)$ starting from a Gaussian noise sample, according to Eq. \ref{eq:reverse}.


\subsection{Conditional Diffusion Sampling} \label{subsec:condsample}


Our key use case for generative models is conditional sampling with one or more constraints. 
When conditionally sampling a diffusion model, one can use {\em classifier-based}~\cite{dhariwal2021diffusion} or {\em classifier-free}~\cite{ho2022classifier} guidance. 
Classifier-based guidance uses the gradient of a classifier to bias the sampling of an
unconditional diffusion model.  
In contrast, the classifier-free guidance doesn't require
another model 
but assumes knowledge of all conditions at training time. 

We hope to achieve compositional generality by allowing the model to work on new tasks when given new constraints. Namely, we don't assume that all potential constraint types are known when training the models, but would like to compose new constraints directly with the existing models. Thus, we opt to use {\em classifier-based} guidance as it allows us to combine the unconditional diffusion model with new classifiers after it is trained.

For classifier guidance, as shown by Dhariwal \emph{et al.}~\cite{dhariwal2021diffusion}, the denoising procedure 
can be approximated as 

{\small
\begin{equation}\label{eq:condition}
    p(x^{(t-1)} \mid x^{(t)}) \approx \mathcal{N}(x^{(t-1)}; \mu_\theta + \Sigma_\theta g_\phi ,\Sigma_\theta),
\end{equation}
}
where $g_\phi = \nabla_{x^{(t)}} \log(p_\phi(y\mid x^{(t)}))$
is the gradient from a classifier $p_\phi(\cdot)$, that models the likelihood of sample $x^{(t)}$ having property $y$, to bias the sampling. This sampling process can be significantly more efficient than rejection sampling if one's goal is to get $x^{(0)}$ that has property $y$. 

\subsection{Latent Parameter Encoding}

In our \tamp{} domain, we are interested in learning samplers that operate on objects that have never seen before, without access to a model of their shape and kinematics.
We can only sense them through
observations $\omega$ in the form of segmented partial point clouds,  
projected from depth images. 
Moreover, for articulated objects, the geometry can non-rigidly change over time. Modeling the explicit dynamics and transitions of such changes in the space of 3D point is challenging and inefficient.

For a more compact encoding and to make the constraint learning problem easier, we train a point cloud encoder $\phi_{enc}$ 
to encode the observation $\omega$ into latent state~$z$. 
Observed partial point cloud $\omega \in \mathcal{R}^{N \times 3}$ is represented as a latent vector~$z \in \mathcal{R}^{d_z}$, where~$d_z$ is the latent dimension. 

\subsection{Examples of Learned Samplers and Classifiers}
\label{sec:push}
In the following examples, we use variables $o$ for object type information, $z$ for latent object shape representations, $q$ for robot configurations, $g$ for a grasp transform, and $p$ the pose of objects of a known shape.



\subsubsection{Pushing Trajectory Sampler} 
\label{subsec:push_sampler}

Non-prehensile motions that involve continual contact such as pushing are difficult to plan for, especially if the object is not known \emph{a priori}. We hope a pushing trajectory sampler can model the constraints and dynamics of a robot interacting with an articulated object via pushing.
\begin{equation}
    p(x) = p(o, z_1, q_1, z_2, q_2, ..., z_T, q_T).
\end{equation}

The vanilla \pddl{DiffPush} sampler models a distribution of trajectories. The trajectory contains both the robot configuration $q$ and latent state $z$ of object $o$, over $T$ time steps. Different from the commonly used diffusion policy, this is a transition model, and not a policy directly.

Recall that we can bias the diffusion sampling, drawing a conditional sample from this trajectory distribution can be done by adding such biases on some of these variables $(z_1, ..., z_T, q_1, ..., q_T)$. We can apply conditionings to different sets of variables, which creates a set of different samplers. If the start of the trajectory $z_1$ is known and we want to predict the states in the following time steps,  we can create a {\em forward} sampler $p(q_1, z_2, q_2, ..., z_T, q_T {\mid} o, z_1)$ by enforcing $z_1 = \mathbf{z_1}$. Similarly, when the end state $z_T$ is given, we can have a {\em backward} sampler $p(z_1, q_1, z_2, q_2, ..., q_T {\mid} o, z_T)$ that infers the {\em pre-image} of latent state $z_T$. Or, if one wants to sample the possible trajectories that can change the object state from $z_1=\mathbf{z_1}$ to $z_T=\mathbf{z_T}$, we can supply a {\em bi-directional} sampler $p(q_1, z_2, q_2, ..., q_{T-1} {\mid} o, z_1, z_T)$ that fills in possible transitions.

In an H-CSP solving context, depending on the variable ordering, the \pddl{DiffPush} sampler can be applied in multiple ways. For example, if the CSP solver samples a target state $z_T$ prior to sampling a pushing trajectory, where $z_T$ needs to satisfy another constraint $c_\psi$, $z_1$ is observed, then the {\em bi-directional} sampler can be used to find a robot-object trajectory
\begin{equation}
    p(q_1, z_2, q_2, ..., q_{T-1} \mid o, z_1, [c_\psi(o, z_T) = \True]).
    \label{eq:pushtrajcondsample}
\end{equation}

\subsubsection{Object State Classifiers}

Often, the goal conditions that define $S_*$ require specific objects to be in a configuration that is semantically meaningful for a human, for example, a state that a human considers door being open \pddl{DoorOpen[$\cdot$]} or closed \pddl{Doorclosed[$\cdot$]} based on a point cloud observation.
To model this, we learn classifiers $c_*(o, z)$ on the object $o$ and the latent state $z$.
These classifiers can be applied on already-sampled object state $z$, to filter invalid samples, as a rejection sampler. Alternatively, with a diffusion sampler, it can be used to bias the sampling while generating conditional samples, as described in Sec.~\ref{subsec:condsample}. 
For example, the gradient from a \pddl{DoorClosed[$\cdot$]} classifier can guide a \pddl{DiffPush} sampler to generate robot and microwave trajectory, that the door ends at a closed state, as shown in Fig.~\ref{fig:teaser}. 

\subsubsection{Collision Classifier}

Similarly, trajectory samples must not collide with other objects, for example, a door being opened should not collide with another object.
To model this, we learn a classifier $c_\zeta(o, z, o_2, p_2)$ for constraint \pddl{PairwiseCollision} on the object $o$ and its latent state $z$ versus another object $o_2$ and its relative pose $p_2$.

\section{Implementation}

We now ground our general approach of \tamp{} using diffusion models as samplers in a concrete domain, as shown in Fig.~\ref{fig:teaser}. We consider a robot manipulating a microwave to achieve goals, by planning using a set of learned models.

\subsection{TAMP Formulation}

We instantiate our TAMP problems $\Pi$ using PDDLStream~\cite{garrett2020PDDLStream}, an extension of Planning Domain Definition Language (PDDL)~\cite{Fox_2003} that supports planning with continuous values using sampling operations. Planning state variables and action constraints are represented using {\em predicates}, which are essentially named classifiers. They are grounded using the classifier introduced in the previous section.

The set of goal states $S_*$ is described by a logical formula over predicate atoms.
For example, a goal for the microwave $o_m$ to be open and an object $o_1$ is stowed in the microwave can be defined by:

{\small
\begin{equation*}
    \begin{split}
        \exists\; z.\; &\pddl{AtLatentState}(o_m, z) \wedge \underline{\pddl{DoorOpen}(o_m, z)} \\&\wedge \pddl{Stowed($o_1$)},
    \end{split}
\end{equation*}
}

where the predicate \pddl{DoorOpen} is a learned classifier.

\subsection{Actions with Learned Constraints}


The \pddl{push} action involves the major learned constraints in this domain (Fig.~\ref{fig:push}).
Its parameters are an articulated object $o$ and a sequence of latent states $z_1, ..., z_T$ and robot configurations $q_1, ..., q_T$.
After applying the action, the robot moved from configuration $q_1 {\to} q_T$ and object $o$ moved from latent state $z_1 {\to} z_T$.
The key constraint is \pddl{DiffPush}. \pddl{DiffPush} can be of different forms depending on the variable ordering while solving the CSP, such as the aforementioned forward sampler and bi-directional sampler. 

\begin{figure}[h]
\begin{small}
\begin{lstlisting}
push|$(o, z_1, q_1, ..., z_T, q_T)$|
  |\kw{con}:| [|\underline{DiffPush$(o, z_1, q_1, ..., z_T, q_T)$}|, 
    |$\neg$\underline{Unsafe$(o, z_1)$}|, ..., |$\neg$\underline{Unsafe$(o, z_T)$}|]
  |\kw{pre}:| [AtLatentState|$(o, z_1)$|, AtRobotConf|$(q_1)$|]
  |\kw{eff}:| [AtLatentState|$(o, z_T)$|, AtRobotConf|$(q_T)$|,
    |$\neg$|AtLatentState|$(o, z_1)$|, |$\neg$|AtRobotConf|$(q_1)$]
\end{lstlisting}
\end{small}
\caption{The \pddl{push} action description.}\label{fig:push}
\end{figure}

$\pddl{Unsafe}(o, z)$ imposes the collision-free constraint on the state $z$ on all other objects, which is evaluated by the learned classifier \pddlsmall{PairwiseCollision}. Pairwise collision is checked between $o$ and all other objects in the domain.

{\small
\begin{align*}
    \pddl{Unsafe}&(o, z) \equiv \exists\; o_2, p_2.\; \pddl{AtPlace}(o_2, p_2) \\
    & \wedge \underline{\pddl{PairwiseCollision}(o, z, o_2, p_2)}.
\end{align*}
}

Consider a problem where the microwave is initially closed, but the goal is for the block to be in the microwave.
The \tamp{} system needs to infer that it should push the microwave sufficiently open so that it can pick and stow the block. If there is obstacle, it also needs to infer that the obstacle should be moved away before opening the door.
Below is an example plan structure. Values in bold are fixed as they are initial state given to the system. 

{\small
\begin{align*}
    \pi = [&\pddl{MoveArm}({\bf q_0}, \tau_1, q_1), \pddl{Push}(o_m, {\bf z_1}, q_1, ..., z_2, q_2), \\
    &\pddl{MoveArm}(q_2, \tau_2, q_3), \pddl{Pick}(\pddl{block}, g, {\bf p_0}, q_3), \\
    &\pddl{MoveArm}(q_3, \tau_3, q_4), \pddl{Place}(\pddl{block}, g, p_*, q_4)]
\end{align*}
}

\begin{figure}
    \begin{subfigure}{\linewidth}
        \begin{subfigure}{0.45\linewidth}
            \centering
            \includegraphics[trim={7.5cm 3cm 14.4cm 2.5cm},clip,width=\linewidth]{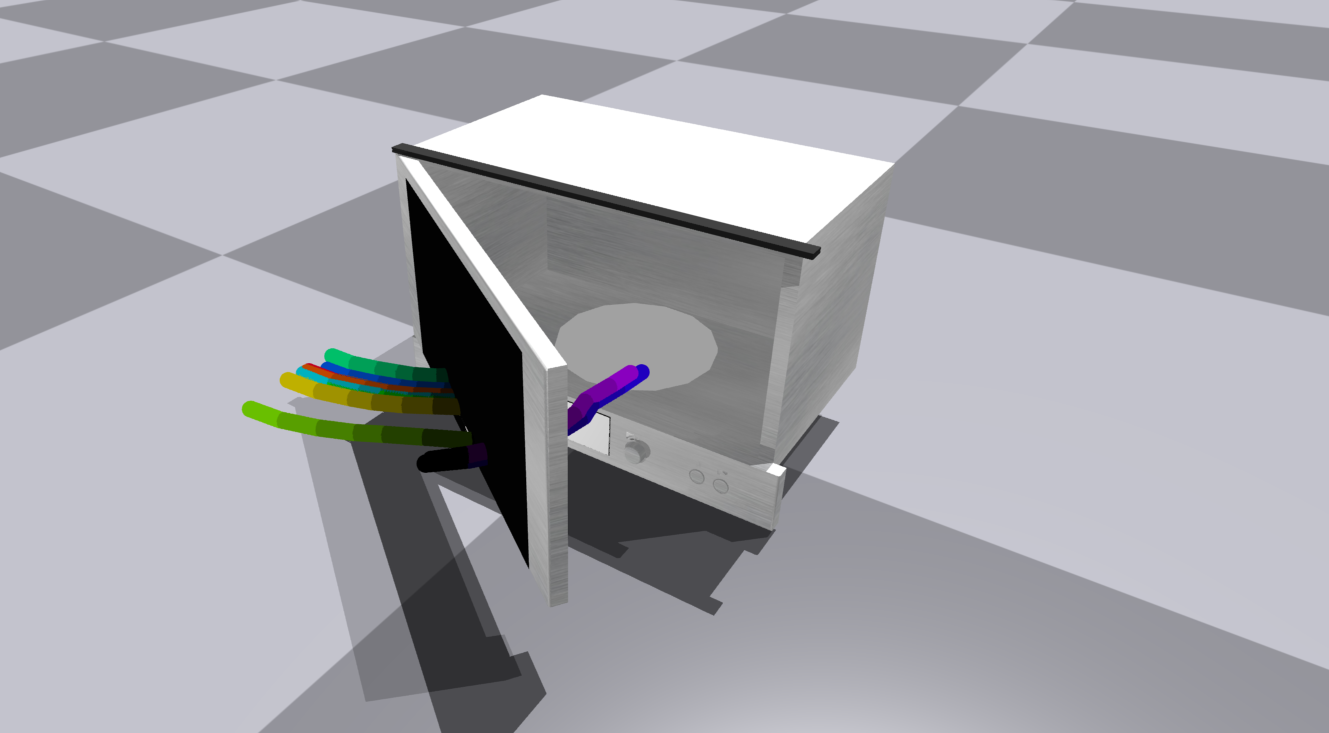}
            \caption{Unconditional}
            \label{}
        \end{subfigure}
        \begin{subfigure}{0.45\linewidth}
            \centering
            \includegraphics[trim={8.5cm 2cm 12cm 4cm},clip,width=\linewidth]{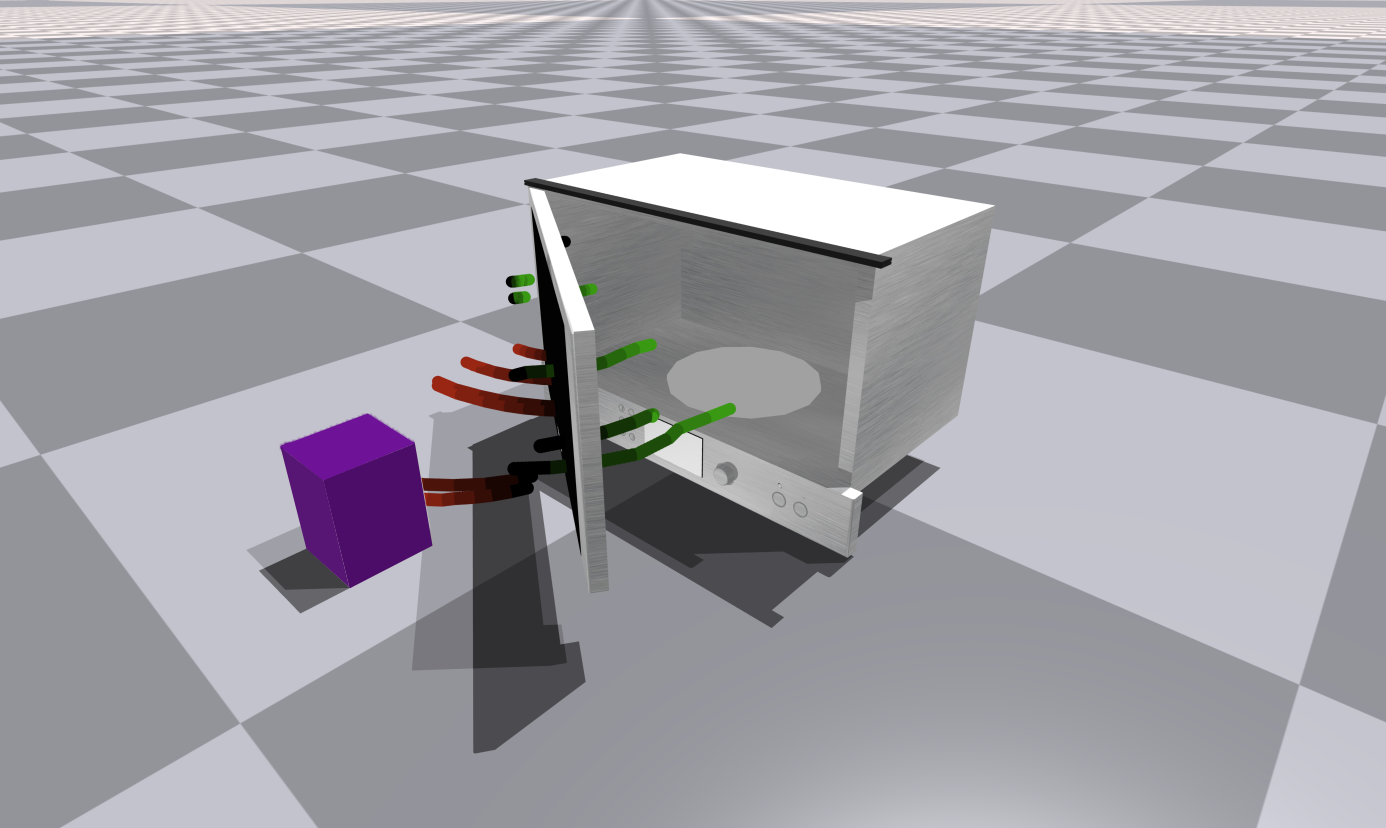}
            \caption{PairwiseCollision}
            \label{}
        \end{subfigure}

    \end{subfigure}
    
    \begin{subfigure}{\linewidth}
        \begin{subfigure}{0.45\linewidth}
            \centering
            \includegraphics[trim={10cm 3.5cm 15cm 4cm},clip,width=\linewidth]{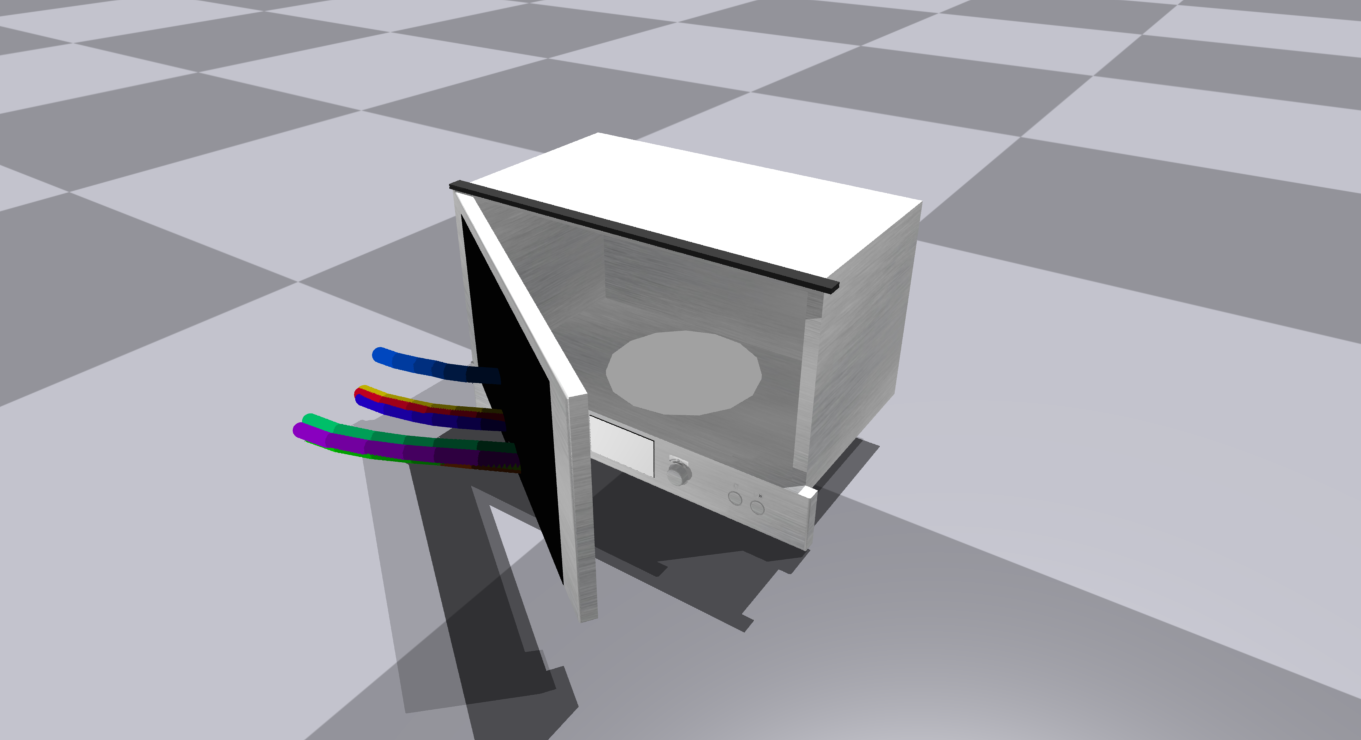}
            \caption{DoorOpen} 
            \label{}
        \end{subfigure}
        \begin{subfigure}{0.45\linewidth}
            \centering
            \includegraphics[trim={8cm 2cm 12cm 4cm},clip,width=\linewidth]{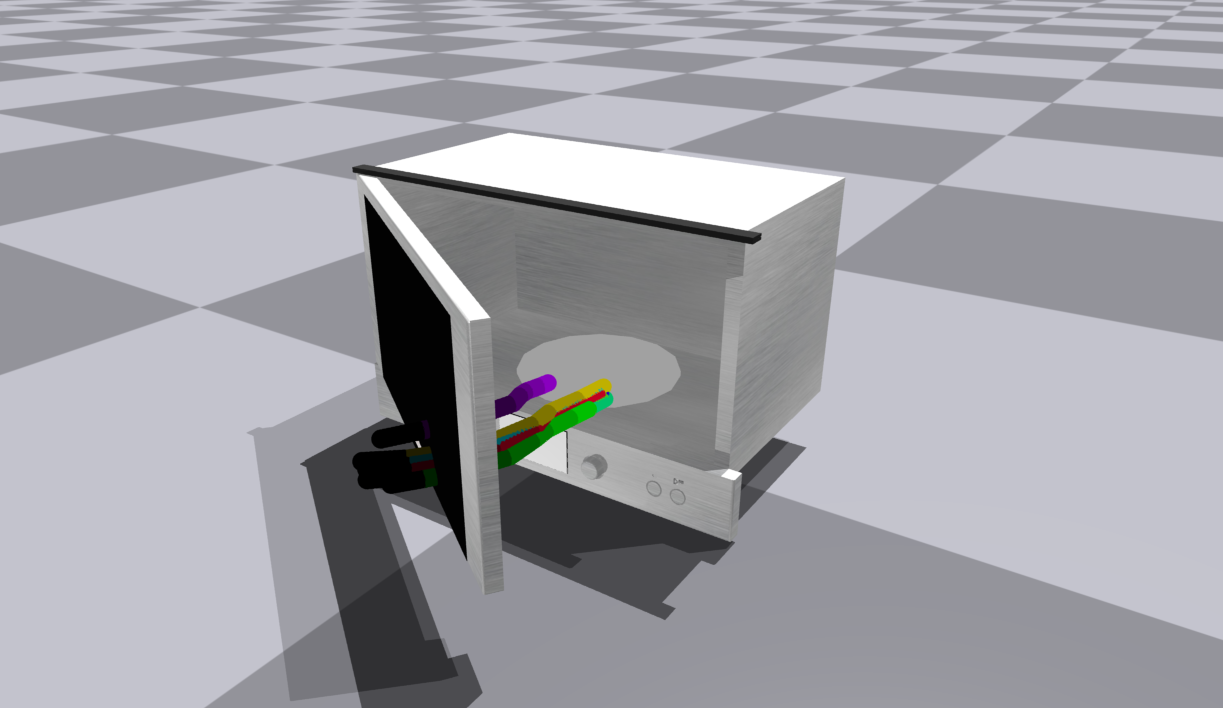}
            \caption{Doorclosed} 
            \label{}
        \end{subfigure}

    \end{subfigure}

    \caption{Samples from the \pddl{DiffPush} model. (a) No condition except for known $z_1$. (b) Checking collision with the purple obstacle using classifier \pddl{PairwiseCollision}. Rejected samples are colored red. (c) (d) Classifier-guided conditional sampling with \pddl{DoorOpen} and \pddl{Doorclosed}. 
    }
    \label{fig:all_traj}
\end{figure}

\subsection{Environment and Data Collection} \label{subsec:environment}


We perform our experiments in the IsaacGym~\cite{makoviychuk2021isaac} physics simulator and verify the learned sampler and classifier in the real world.
To learn the data distribution that satisfies \pddl{DiffPush} constraint,
we generate a training set $\mathcal{D}$ of push action instances by simulating manipulation trajectories in IsaacGym. The manipulation trajectories are generated using a script policy that has access to the ground truth state, with known object models and contact forces. An action sequence consists of end-effector waypoints pushing perpendicular to the segmented door surface. In the training set, we use 10 microwave assets from the PartNet~\cite{mo2019partnet} dataset. We capture the depth images from randomized viewpoints. There are 101 valid trajectories in total, which are later randomly clipped into shorter segments for training.

During data collection, we treat the robot as a disembodied gripper, the robot's configuration is an end-effector position $q \in \mathbb{R}^3$. This representation yields a simple parameterization of the skill and allows the learned model to be flexibly 
integrated into full arm motion using inverse kinematics and motion planner in a \tamp{} framework. Our real-world experiments demonstrate such flexibility by composing the same learned model with robot grippers that have different geometries.

\subsection{Training Details} \label{subsec:training}

To train the point cloud encoder $\phi_{enc}$, we use the aforementioned trajectory dataset, which contains 4746 point clouds. We use PointNet++~\cite{qi2017pointnet++} as the encoder network $\phi_{enc}$ and use the same decoder $\phi_{dec}$ as Cai \emph{et al.}~\cite{cai2020ShapeGF}. The latent vector size $d_z$ is 256. The network is trained with a batch size of 196. All other hyper-parameters are the same as in Cai \emph{et al.}~\cite{cai2020ShapeGF}. After the point cloud encoder is trained, the weights are frozen and used as the encoder when training all other models.

For the trajectory diffusion model, we use a U-Net structure similar to that in Diffuser~\cite{janner2022diffuser}, with 250 diffusion steps, trajectory length 8, and batch size 32. Microwave state (open/close) classifiers are 3-layer MLPs. The inputs to the collision network \pddl{PairwiseCollision} are latent point cloud embedding, obstacle location, and obstacle size approximated as bounding boxes. 

\section{Experiments} \label{sec:experiments}

We carry out standalone evaluations on our learned samplers, and further evaluate them in the integrated \tamp{} setting with joint searching and sampling. The evaluations are done in IsaacGym simulator. We also apply our learned sampler on a real robot, as described in section~\ref{sec:realrobot}.

\subsection{Generating Samples with Specific Constraint}

To verify that our learned push sampler can generate valid and accurate samples, we impose tight constraints on the initial and target state, and exploit the \textit{bi-directional} sampler to fill in intermediate steps in trajectories.
$$p(q_1, z_2, q_2, ..., q_{T-1} {\mid} o, z_1, z_T)$$
We compare with other types of non-diffusion-based models and report the error on door angle after executing the sampled robot trajectory $[q_1, \ldots, q_{T-1}]$ in simulation.


The initial and target microwave door angles are randomly sampled. We capture the point cloud at such configurations $\omega_1, \omega_T$ from a random viewpoint. The point cloud observations are encoded into latent $z_1, z_T$ and used as conditions to sample the rest of the trajectory from the diffusion model. 

We compared the performance of our model to two baselines: 1) a {\em discriminative model} that directly regresses from a pair of start and end positions $z_1, z_T$ to $q_1, ..., q_{T-1}$ and 2) an energy-based model (EBM).

\begin{table}[t]
    \centering
    \begin{tabular}{l|c}
        \bf{Method} & \bf{Mean Error (Stderr)}\\
        \hline
        Regression model  &  0.127 (0.021) \\
        EBM & 0.191 (0.021)\\
       \bf{Diffusion} & \textbf{0.094 (0.009)}
    \end{tabular}
    \caption{Mean error to the target angle (in radians) after executing the trajectory in simulation. Our method (Diffusion) outperforms the two baselines.} 
    \label{tab:angle_error}
\end{table}

Table~\ref{tab:angle_error} shows the mean (and standard error) of the absolute distance between the target angle and the actual ending angle on 100 testing trajectories. Results are shown in radians.
The diffusion model has the lowest error. Compared to EBM, the diffusion models are easier to train and have a more stable gradient, which is also observed in \cite{chi2023diffusionpolicy}. Though the regression model is also stable during training, the result is deterministic given the same point cloud embeddings $z_1, z_T$, which is not desirable as a sampler according to our analysis in Sec.~\ref{subsec:condsampledesiderata}. In contrast, the diffusion model can generate more diverse samples that satisfy the given constraints. See Fig.~\ref{fig:all_traj} for an illustration. 

\subsection{Generating Conditional Samples with Classifier Guidance}
\label{subsec:abstractgoal}

The previous section evaluates our model on a given goal state $z_T$. We want to verify that our model can also generate trajectories conditioned on a more abstract goal,
$$p(o, q_1, z_2, q_2, ..., q_{T-1} \mid o, z_1, [c_\psi(o, z_T) = \True])$$where the target ending state is given in the form of a constraint. Any trajectory $q_1, ..., q_{T-1}$ leading to a state $z_T$ such that $c_\psi(o, z_T) = \True$ is a constraint-satisfying sample.

\begin{figure}
    \centering
    \begin{subfigure}{\linewidth}
    \centering
        \begin{subfigure}{.4\linewidth}
        \centering
            \includegraphics[trim={0cm 2cm 2cm 2.3cm}, clip, width=\linewidth]{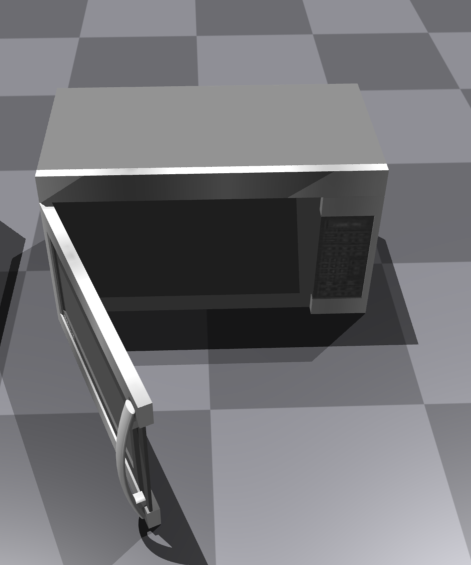}
        \caption{Initial microwave state.}
        \end{subfigure}
        \begin{subfigure}{.58\linewidth}
        \centering
            \fbox{\includegraphics[trim={2cm 2cm 2.5cm 2cm}, clip, width=\linewidth]{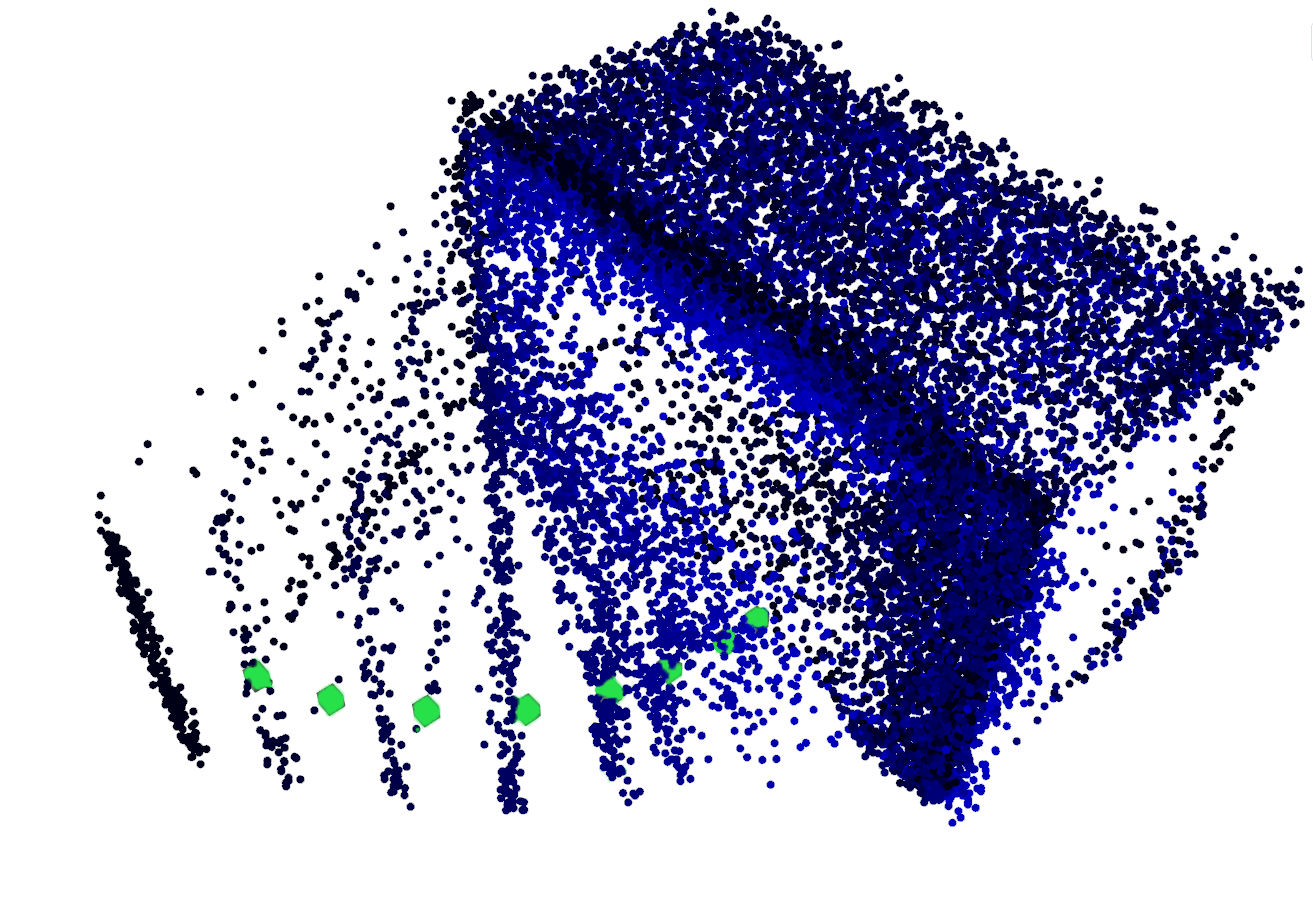}}
        \caption{Predicted microwave trajectory.}
        \label{fig:state_prediction_traj}
        \end{subfigure}
    \end{subfigure}

    \caption{One sampled trajectory of the microwave from the initial state (left figure) to the ``fully closed'' state. Sampled latent state $z$ is decoded into a point cloud and rotated for better visualization.}
    \label{fig:state_prediction}
    \vspace{-5pt}
\end{figure}

We evaluate the learned model with classifier-guided conditional sampling, where $c_\psi$ is given in the form of named classifiers, according to Eq.~\ref{eq:condition}. We train three classifiers to model the point cloud as being ``fully closed'', ``fully open'', ``half closed'', and ``half open''. The corresponding semantics are door angle less than 0.2, greater than 1.4, less than 1.2, and greater than 1.2 radians. 
We evaluate 100 random initial configurations of the microwave and report the success rate of the ending state after executing the sampled trajectory $q_1, ..., q_{T-1}$ (Table~\ref{tab:abstractgoal}). Qualitative results are shown in Fig.~\ref{fig:all_traj}, with the door being fully closed and fully opened. 
Fig.~\ref{fig:state_prediction} shows the entire sampled trajectory with classifier guidance of ``fully closed''. The green boxes show the action waypoints $q_1, ..., q_{T-1}$.
\begin{table}[h!]
    \centering
    \begin{tabular}{l|c|c|c|c}
       \bf{Goal} & \bf{Fully Closed} & \bf{Fully Open} &  \bf{Half Closed} & \bf{Half Open} \\
       \hline  
       Success & 94\% & 92\% & 100\% & 90\%
    \end{tabular}
    \caption{Success rates for achieving abstract goals using classifier guidance.}
    \label{tab:abstractgoal}
\end{table}

Our model achieves a high success rate on all semantic goals. Further, the classifiers are trained completely independent of the diffusion trajectory sampler. This is valuable for generalization to new domains with novel constraint types. One can re-use the unconditional trajectory model and only learn the classifier that encodes the semantics of the new constraint.

\subsection{\tamp{}: Joint searching and sampling} 
\label{subsec:obstacletask}
In prior experiments, we verify the sampling capability of our learned models. In the following section, we test the learned models in a complete \tamp{} system. Diffusion models and learned classifiers are combined with other samplers such as IK and motion planners, in order to solve a multi-step manipulation task. The end-effector in previous experiments is simplified as a moving gripper. In this experiment, we add the full-arm motion and enforce additional constraints such as the sampled end-effector poses being reachable. These additional constraints need to be considered jointly. If none of the sampled trajectories satisfy all constraints, we either draw more samples, or search for a new task plan that has a different constraint relation.
We use PDDLStream to jointly search the task plan and make calls to corresponding samplers.

\begin{figure}[t]
        \centering
        \begin{subfigure}{.46\linewidth}
        \centering
            \includegraphics[trim={10cm 7cm 12cm 5cm}, clip, width=\linewidth]{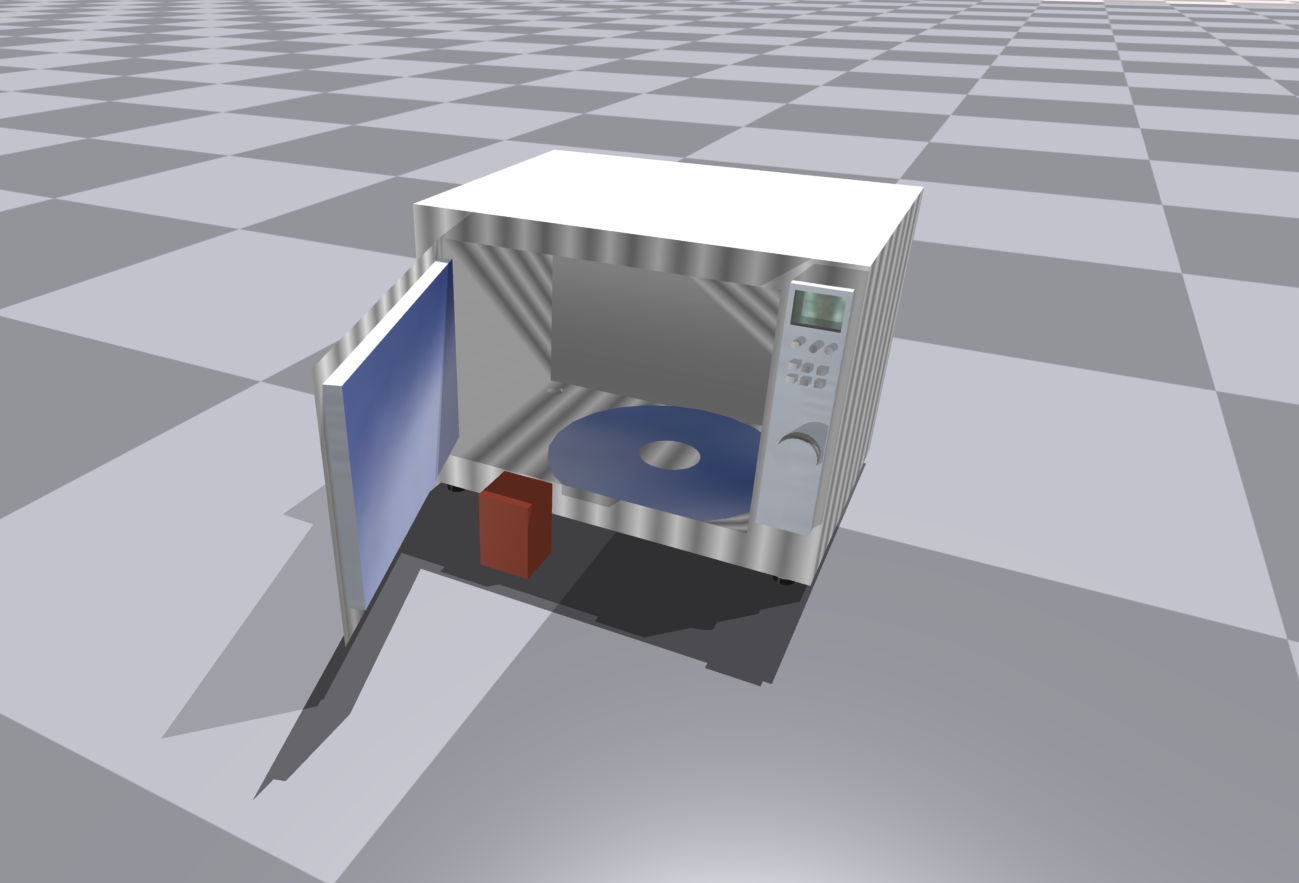}
        \caption{The \textit{Close} task.}
        \label{}
        \end{subfigure}
        \begin{subfigure}{.42\linewidth}
        \centering
            \includegraphics[trim={10cm 5.5cm 8cm 1cm}, clip, width=\linewidth]{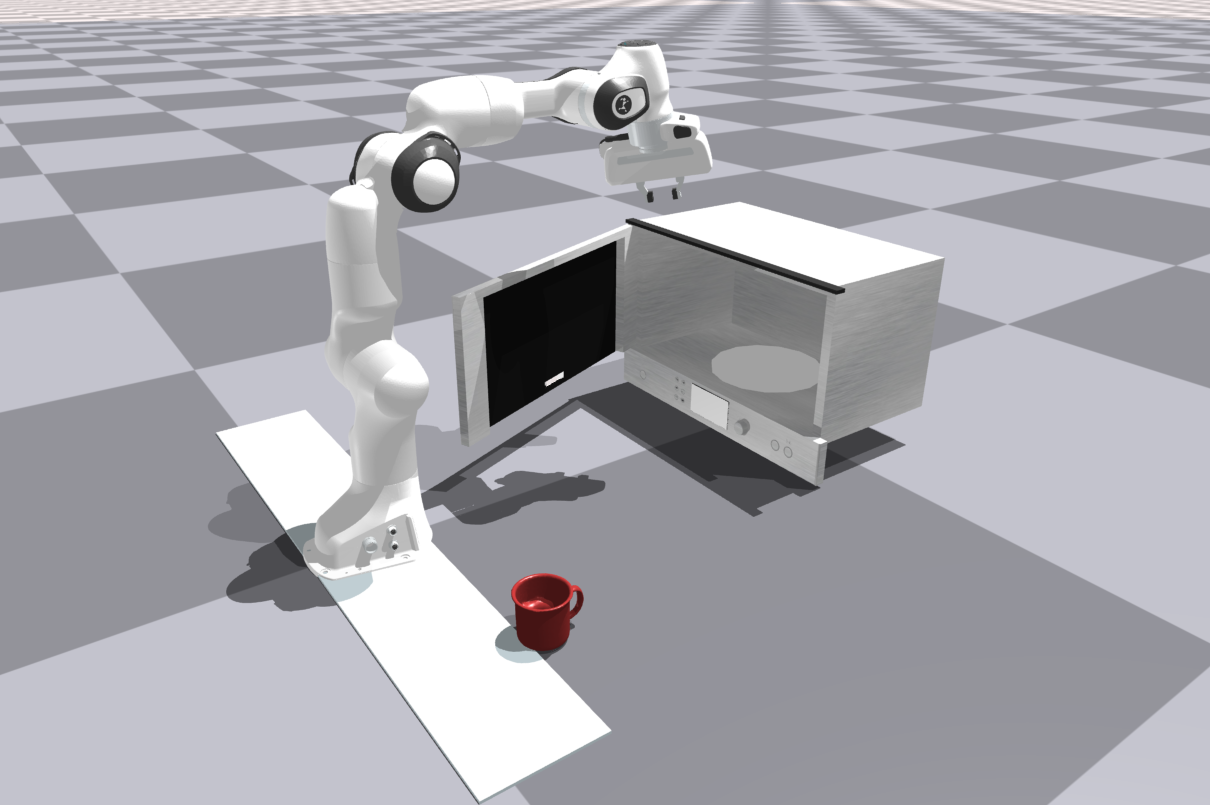}
        \caption{The \textit{Stow-Close} task.}
        \end{subfigure}\\[1ex]
        \begin{subfigure}{.5\linewidth}
        \centering
            \includegraphics[trim={5cm 7cm 12cm 2cm}, clip, width=.9\linewidth]{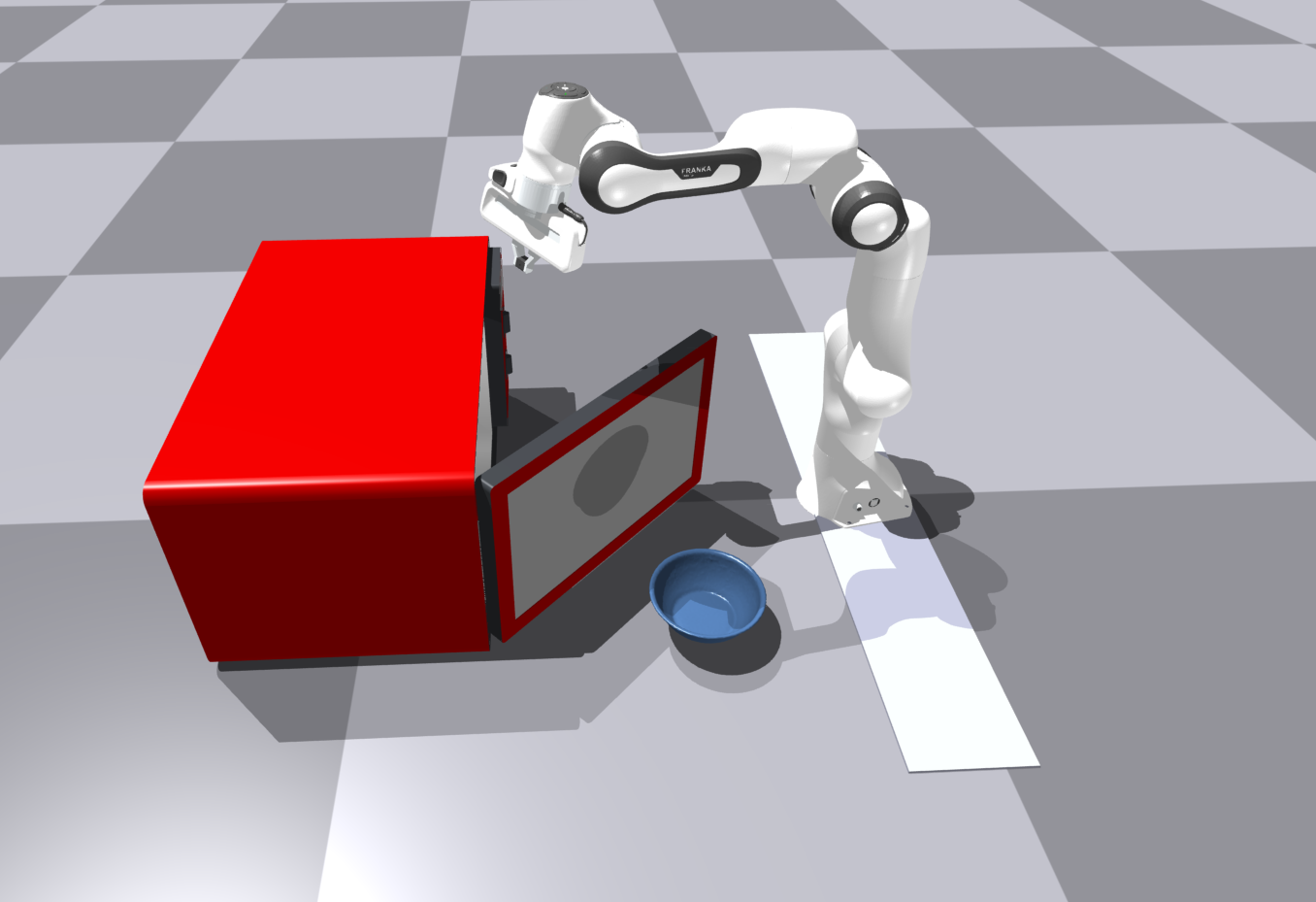}
        \caption{The \textit{Stow-Close-B} task.}
        \label{fig:tamptask_stowingwobstacle}
        \end{subfigure}
    \caption{The initial states in the (a) \textit{Close}, (b) \textit{Stow-Close}, (c) \textit{Stow-Close-B} simulated tasks.} 
    \label{fig:pddlstream_task}
\end{figure}

We evaluate three tasks, shown in Fig.~\ref{fig:pddlstream_task}. 1) \textit{Close}: The goal state is the microwave door at the closed location. The initial state is the door at an opened position, where there is an obstacle blocking the door from directly being closed. 2) \textit{Stow-Close}: The goal is to have an object stowed in the microwave and the door is closed at the end. The object is initialized to be at a fixed location that won't block the microwave. 3) \textit{Stow-Close-B}: The goal is the same as \textit{Stow-Close}. However, the door needs to be fully opened before stowing, as the object is larger. The object is initialized near the microwave which may block the opening action.
As in the previous experiments, the planner only receives a partial point cloud of the microwave captured from a random viewpoint. The initial location and geometry of the object to be stowed and the obstacle are known. We use a Franka Emika robot arm with a mobile base link along the X-axis. The predicted waypoints $q_1, ..., q_{T-1}$ are set as the target endpoint of the gripper. We use the robot URDF to compute IK solutions for the arm. For collision checking between the robot arm and a predicted microwave state $z$, we approximate links of the arm using bounding boxes and query the learned collision checker. We use the learned latent collision checker for all predicted states and the partial point cloud for the initial state.
We compute the motion plans using a bidirectional RRT.

We conduct an ablation on rejection sampling and classifier-guided conditional sampling. Rejection sampling refers to the variant that uses a {\em forward} sampler $p(q_1, z_2, q_2, ..., z_T, q_T {\mid} o, z_1)$ and testing post-hoc whether $c_\psi(o, z_T)=\True$. Conditional sampling refers to using the goal-conditioned sampler as used in Sec.~\ref{subsec:abstractgoal}

We use the \textit{Adaptive} algorithm in PDDLStream with a \textit{search-sample} ratio of 1/15 for all tasks. We use weighted A$^{*}$ and fast-forward heuristic. 
Planning success rates~(PSR) and execution success rates (ESR) are shown in Table \ref{tab:pddlstreamtest}. PSR refers to the success rate during planning - the ratio of problems solved by the planner (which does not necessarily lead to successful execution). ESR refers to that when executing a trajectory, the resulting state actually achieves the goal, among the ones solved by the planner. We evaluate each task for 15 runs. Average high-level action numbers are reported based on solved runs. The high-level actions the planner considers include \pddl{Move($\cdot$)}, \pddl{Stow($\cdot$)}, \pddl{Push($\cdot$)}. Even without considering which object to apply the action, there are 3 possible combinations for plan length 1 and $3^4$ combinations for a plan with 4 actions.
Examples of the final plans are available at the \href{https://sites.google.com/view/dimsam-tamp}{website}.

\begin{table}[t]
    \centering
    \begin{tabular}{l|r|r|r}
    \toprule
        \bf{Task}  & \bf{\#Actions} & \bf{PSR} &  \bf{ESR} \\
    \hline
    {Close} & 1.93 & 1.00 & 0.60 \\ 
    {Close$^{c*}$} & 1.93 & 1.00 & 0.87 \\ 
    {Stow-Close} & 1.93 & 0.93  & 0.71 \\ %
    {Stow-Close$^{c*}$}  & 2.00 & 1.00 &  1.00 \\ %
    {Stow-Close-B$^{c*}$} & 4.00  & 0.87 & 1.00 \\
    \bottomrule
    \end{tabular}
    \caption{Results on joint planning using PDDLStream. Task names with the superscript $\cdot^{c*}$ use conditional samplers. PSR stands for planning success rates. ESR stands for execution success rates.}
    \label{tab:pddlstreamtest}
\end{table}

\textbf{Behavior Analysis} 
In the first task, the first plan found by the planner will 
try to close the door directly \pddl{[Push$(o_m)$]}. Constraints in this plan include [{\small \pddl{DoorClosed($o_m, z_T$)}, \pddl{DiffPush($o_m, z_1,q_1,\ldots,z_T$)}, \pddl{$\neg$Unsafe$(o_m, z_*)$}}], etc. These constraints are not satisfiable, since there is an obstacle blocking the door from closing, so there is no collision-free path for the microwave to close, without moving the blocking object first (\pddl{$\neg$Unsafe$(o_m, z_*)$} is unsatisfiable).
With returned failures from the samplers, the planner will start searching for an alternative plan.
The most typical satisfiable plan found is \pddl{[Move$(o_1)$, Push$(o_m)$]}. We found the classifier-guided sampling Close$^{c*}$ to have a higher execution success rate, as guided sampling of the diffusion model (as in Eq.~\ref{eq:pushtrajcondsample}) tends to generate higher quality samples than the \textit{forward} sampler.
In \textit{Stow-Close}, the planner needs to reason about the order to achieve the subgoals in \pddl{[Push$(o_m)$, Stow$(o_1)$]}. Closing the door prior to stowing object can achieve one subgoal, but making the sampling of another subgoal unsatisfiable, without undoing the first action. The \tamp{} planner can reason about such constraints and still has a high PSR. We also have the same observation here that the ESR of using conditional sampler is higher than that of the rejection sampling.
The third task is the most challenging one. A precondition of the door being opened is added to the domain. Due to the combinatorial complexity, the search space grows exponentially larger compared to the previous two tasks. Two out of 15 tests does not return a valid plan.

Among the execution failure modes, we have observed that the sampled trajectory may be too close to the joint axis of the door and making it physically infeasible. We will consider incorporating uncertainty or effort estimates from learned samplers in future work.

It is important to note that this is a problem that cannot be addressed by classical TAMP methods, because the kinematic and shape models of the microwave are unknown.  In addition, note that no new learning was required to solve this problem---once the individual generative models are trained for each constraint, they can be combinatorially recombined to solve a wide variety of problems.   This is in contrast to direct policy learning methods, which require training on new tasks and generally work poorly on multi-step problems that require geometric and constraint reasoning unless given a carefully crafted reward function.

\section{Real World Experiment}\label{sec:realrobot}

\begin{figure}
    \centering
    \begin{subfigure}{\linewidth}
        \begin{subfigure}{0.49\linewidth}
            \centering
            \includegraphics[trim={0 2cm 20cm 4cm},clip,height=3cm]{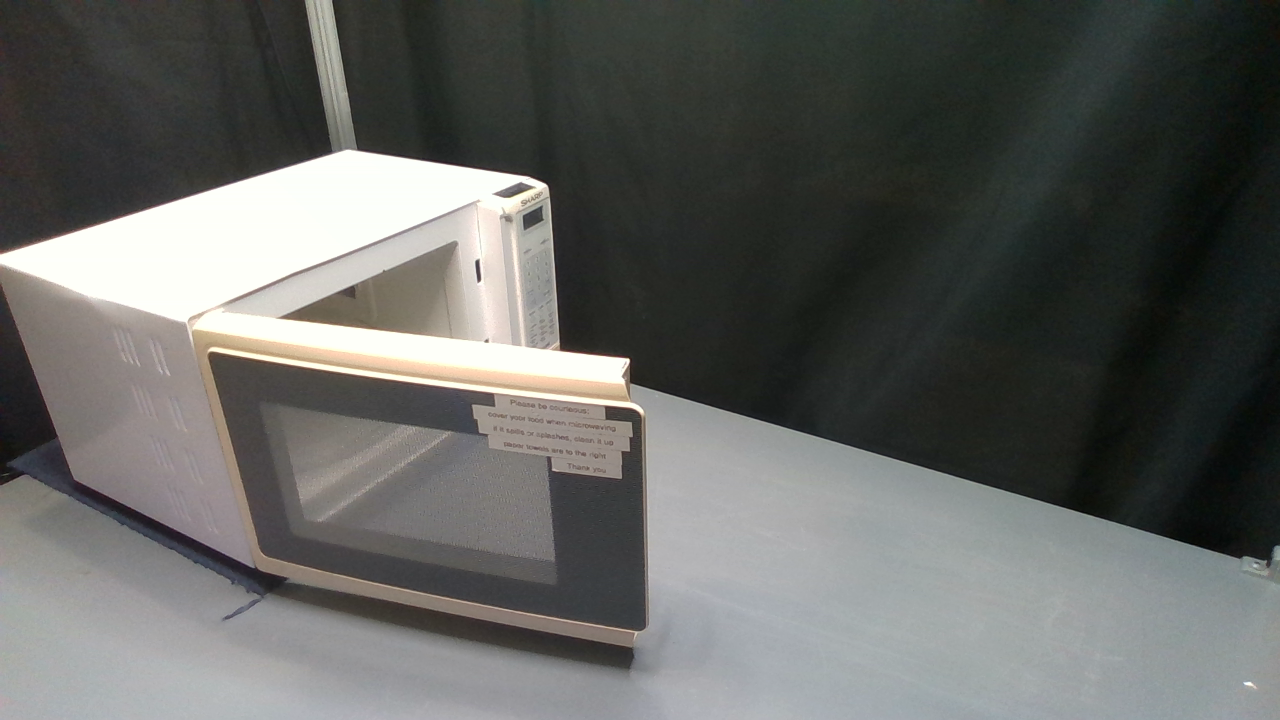}
            \caption{Observed RGB image. Models only use the depth data.} 
            \label{}
        \end{subfigure}
        \begin{subfigure}{0.49\linewidth}
            \centering
            \fbox{\includegraphics[trim={10cm 0 8cm 0},clip,height=3cm]{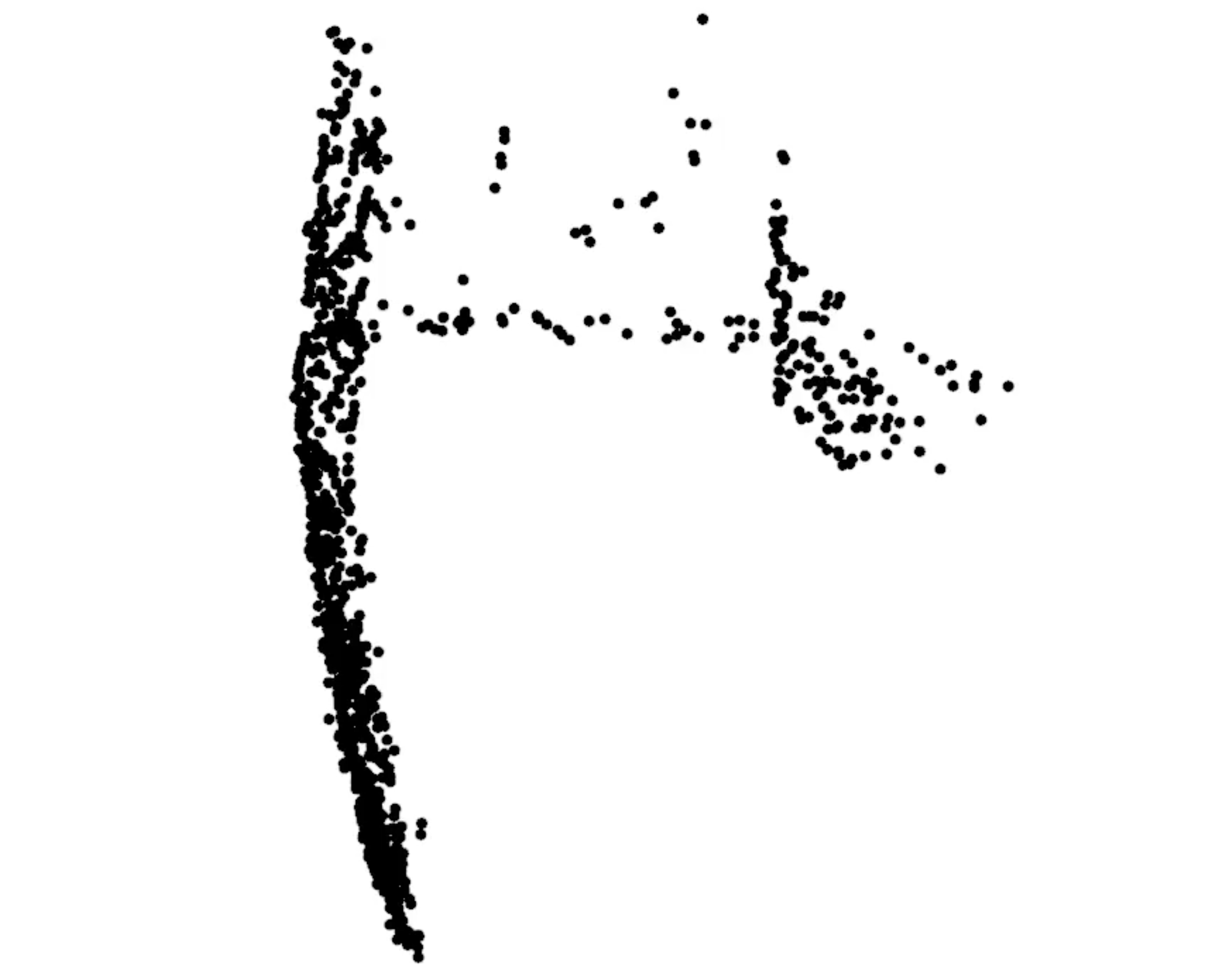}}
            \caption{Observed point cloud. Rotated to top-down for better visualization.}
            \label{fig:observe_realdep}
        \end{subfigure}
    \end{subfigure}
    \begin{subfigure}{\linewidth}
        \begin{subfigure}{0.49\linewidth}
            \centering
            \includegraphics[trim={0 0 0 0},clip,width=\linewidth]{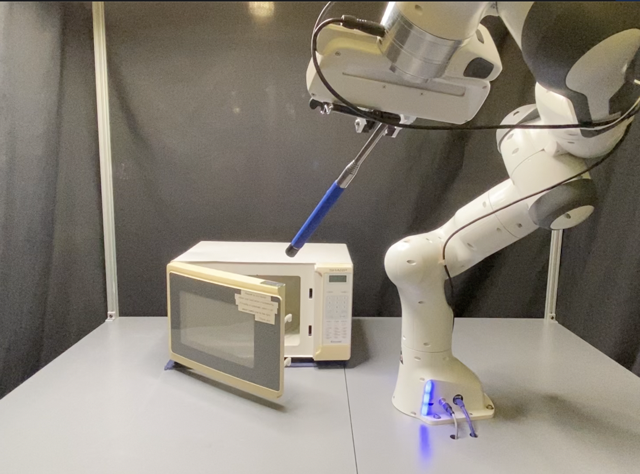}
            \caption{The close door task.} 
            \label{}
        \end{subfigure}
        \begin{subfigure}{0.49\linewidth}
            \centering
            \includegraphics[trim={0 0 0 0},clip,width=\linewidth]{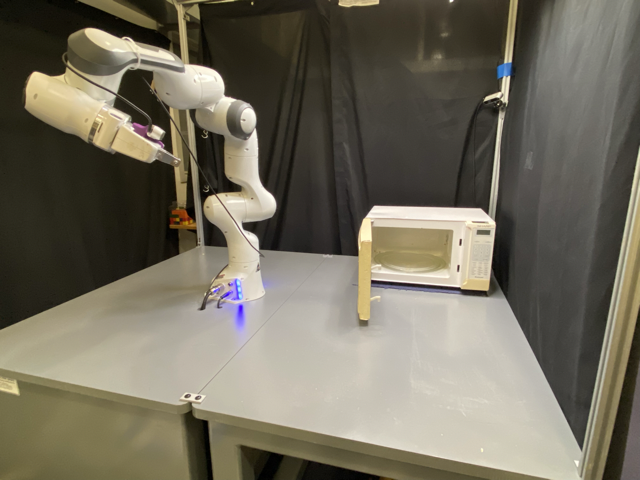}
            \caption{The open door task.} 
            \label{}
        \end{subfigure}
    \end{subfigure}

    \caption{Real world experiment setup and model observation.}
    \label{fig:realworld}
\end{figure}

We apply the model trained in simulation directly to the real world, without finetuning, on door closing and opening tasks.
We capture the depth image from a RealSense D435i depth camera mounted on the Franka gripper. The point cloud of the microwave is segmented by removing the table plane. We apply the learned \pddl{DiffPush} model trained in simulation on the observed point cloud directly without any finetuning, demonstrating zero-shot sim-to-real transfer. In the door closing and opening tasks, we use conditional sampling based on \pddl{Doorclosed} and \pddl{DoorOpen} classifiers trained in simulation. The planner searches for trajectories that satisfy reachability and collision-free constraints.

In the door-opening task, 
the Franka robot gripper is too wide to fit in between the microwave door and the microwave base. We augment the robot with a tool that it can use to make contact with the microwave. The waypoints predicted by our diffusion models are set as the target locations of the tool endpoint. Since the action representation in the learned model is simplified as waypoints, we don't need to re-train the model but only need to modify the IK component.

The observed RGB image and point cloud are shown in Fig.~\ref{fig:realworld}. The RGB image itself is for illustration only and is not used by the model. Due to the reflective material of the microwave and the glass used in the microwave door, there are a lot of missing depth readings, as shown in Fig.~\ref{fig:observe_realdep}. Despite this severe partial observability, our model is still able to generate actionable target samples. Videos showing the robot solving these tasks are available at: \href{https://sites.google.com/view/dimsam-tamp}{https://sites.google.com/view/dimsam-tamp}. Thanks to the modular design of the \tamp{} system and the choice of observation and action space in learned models, we are able to deploy the system in the real world without finetuning.


\section{Conclusion}
We use diffusion models for sampler learning and compose
the learned samplers in a TAMP framework. Importantly, our framework can {\em both} reason about the set of constraints that need to be satisfied, and use the learned models in drawing constraint-satisfying samples. 
We have showed that conditional sampling on learned models is efficient and effective.
We instantiated an example of such samplers in an articulation manipulation domain through learning pushing constraints. Finally, we compared our diffusion approach with several baselines in simulation, demonstrated it within
a TAMP system for completing multi-stage tasks, and ultimately deployed it in the real-world. We believe this strategy can be applied in broader domains.

\section{Acknowledgement}
We gratefully acknowledge support from NSF grant 2214177; from AFOSR grant FA9550-22-1-0249; from ONR MURI grant N00014-22-1-2740; from ARO grant W911NF-23-1-0034; from the MIT Quest for Intelligence; and from the Boston Dynamics Artificial Intelligence Institute.

\bibliographystyle{IEEEtran}
\bibliography{references}
\end{document}